%% file: paper.tex
\documentclass[letterpaper]{article}

\usepackage{natbib,alifeconf}  

\usepackage{amssymb} 
\usepackage{amsmath} 
\usepackage{xcolor}

\title{Decomposing the Prediction Problem; Autonomous Navigation by neoRL Agents}
\author{Per RL \\
\mbox{}\\
UiT -- Norges Artigste Universtet\\
Per.R.Leikanger@uit.no} 

\begin{document}   
\maketitle


\input{src/abstract}

\input{src/introduction}

\input{src/theory}
\input{src/methods}

\input{src/experiments}
\input{src/discussion}

\footnotesize
\bibliographystyle{apalike}
\bibliography{peerlearning}
\end{document}

%% file: src/abstract.tex
\begin{abstract}
    Navigating the world is a fundamental ability for any living entity.
    Accomplishing the same degree of freedom in technology has proven to be difficult.
    The brain is the only known mechanism capable of voluntary navigation, making neuroscience our best source of inspiration toward autonomy. 
    Assuming that state representation is key, we explore the difference in how the brain and the machine represent the navigational state.
    Where Reinforcement Learning (RL) requires a monolithic state representation in accordance with the Markov property,
        Neural Representation of Euclidean Space (NRES) reflects navigational state via distributed activation patterns.
    We show how NRES-Oriented RL (neoRL) agents are possible before verifying our theoretical findings by experiments.
    Ultimately, neoRL agents are capable of behavior synthesis across state spaces -- allowing for decomposition of the problem into smaller spaces, alleviating the curse of dimensionality.
\end{abstract}

%% file: src/introduction.tex
\section{Introduction}

Autonomy or any form of self-governed activity implies an ability to adapt with experience;
    hard-coded algorithms, agents governed by external control, or deterministic model-based path planning can hardly be said to be autonomous.
``Navigation can be defined as the ability to plan and execute a goal-directed path'' \citep{solstad2009neural}.
Robot motion planning can be defined in similar terms \citep{latombe2012robot}; 
    however, cybernetics and robot motion control involves model with limited validity intervals or algorithms for deterministic control.
The reward hypothesis from Reinforcement Learning (RL) is relevant in this context: 
\emph{``That all of what we mean by goals and purposes can be well thought of as maximization of the expected value of the cumulative sum of a received scalar signal (reward).'' \citep{sutton2018reinforcement}.}
With a proven track record for learning to solve digital challenges or for intelligent games,
    RL agents have demonstrated a capability of autonomy for specific challenges.
Via methods from function approximation by Deep Learning, methods from RL can form agents with superhuman abilities for certain board games \citep{tesauro1994td, silver2016mastering, silver2017masteringChess} and games of hazard \citep{heinrich2016deep}.
However, RL supported by deep function approximation is known to require a tremendous amount of training: 
Robot autonomy by RL remains an unsolved challenge, partially due to requirements for real-time execution and model-uncertainty -- limiting the number of accurate samples for training \citep{kober2013reinforcement}.
RL agents supported by deep function approximation can learn impressive abilities, 
    but statistical machine learning approaches require much experience, do not generalize well, and are monolithic during training and execution \citep{kaelbling2020foundation}. 

Autonomous navigation is an ability unique to the central nervous systems in the animal and insects. 
Determining one's parameter configuration relative to an external reference, one's \emph{allocentric} coordinate, is critical for navigation learning \citep{whitlock2008navigating}.
Several mechanisms have been identified in the brain that represent Euclidean coordinates at the single-neuron level \citep{bicanski2020neuronal}.
Notable examples for navigation are Object Vector Cells \citep{hoydal2020}, representing the allocentric location of objects around the animal, Head-Direction Cells \citep{taube1990head}, representing the heading of the animal, and border cells \citep{solstad2009neural}, representing the proximity of borders for navigation.
Possibly the most well-known cell for Neural Representation of Euclidean Space (NRES) is the \emph{Place Cell}.
This first identified NRES modality represents the allocentric location of the animal \citep{o1971hippocampus}:
    When an animal's location is within the \emph{receptive field} of one place cell, the neuron is active in terms of having a heightened firing frequiency. 
The activation pattern in an appropriate population of NRES neurons can thus map any position in a finite Euclidean space \citep{fyhn2004spatial}.
Other NRES modalities have later been identified, with a similar mechanism for representing coordinates in other Euclidean spaces \citep{bicanski2020neuronal}.
With our sense of orientation originating from multiple NRES modalities, distributed representation of state appears to be of critical importance for navigational autonomy.

This article starts out by presenting important considerations from RL and directions that could allow for a distributed representation of state. 
Off-policy learning allows agents to learn general value functions for independent aspects of a task \citep{sutton2011horde}.
When a hoard of learners base their value function on a mutually exclusive reward signal, inspired by NRES cells, we propose a method for learning an orthogonal basis for behavior.
Experiments with NRES-Oriented RL (neoRL) agents by the Place Cell NRES modality demonstrate how the proposed framework allows for reactive navigation in real-time.

%% file: src/theory.tex

\section{Interaction learning by RL in AI}
Reinforcement learning is the direction in machine learning concerning learning behavior through interaction with an environment.  
We say that the decision \emph{agent} learns to achieve a task according to a scalar reward signal $\mathbb{R}$ by interaction with an \emph{environment}.
The accumulated experience takes the form of agent \emph{value function}, reflecting the benefit of visiting different states or state-actions pairs according to the \emph{reward signal} during training.
When the algorithm learns the value of state-action pairs, i.e., learning the value of selecting specific actions from different states, this is referred to as Q-learning.
An important aspect of RL environments is the \emph{Markov property}: When a state-action pair uniquely defines the probability distribution of the next state, the decision process is referred to as a Markov Decision Process (MDP).
When a problem can be represented as an MDP, 
    an RL-agent can, in theory, learn an optimal solution to tasks expressed by a reward function from interaction alone 
\citep{sutton2018reinforcement}.

The \emph{prediction problem} in reinforcement learning concerns estimating the value of visiting different states $s$ while following policy $\pi$.
The agent state is a compact representation of the history and necessary information for the agent to make a decision at \mbox{time $t$}.
The value function can be updated according to the Bellman equation: 
\begin{equation}
\label{eqBellmanSatateValue}
  v_{\pi}(s) = \sum_a \pi(a|s) \sum_{s', r} p(s',r|s,a)\left[ r+\gamma v_\pi(s') \right]
\end{equation}
Updating the value function under policy $\pi$ from experience gathered while following the policy $\pi$, is referred to as on-policy learning \citep{sutton2018reinforcement}.
Off-policy learning allows an agent to form the value function while following another behavior policy.
Through off-policy learning, an agent can learn the value function under a target policy $\pi_t$ while following a different behavior policy $\pi_b \neq \pi_t$.
The agent can, for example, initially follow a more exploratory policy or learn while observing human control \citep{abbeel2007application}.
Learning the value function is possible through pure observation.

General Value Function (GVF) is one identified use of off-policy learning, where the agent learns value functions potentially unrelated to the control problem \citep{sutton2011horde}.
These partial agents, only concerned with accumulating experience, can be seen as independent \emph{learners} of an auxiliary value function used to answer questions about the environment.
Examples of questions, as listed in the original paper, could be time-to-obstacle or time-to-stop for the Critterbot demonstration \citep{sutton2011horde}.
Auxiliary value functions can also be directly involved in policy, as demonstrated for the Atari game Ms. PacMan. 
A set of General Value Functions were trained for manually designed sub-challenges in the Ms. Packman computer game, 
    resulting in an exponential breakdown of problem size compared to ``single-headed'' RL agents \citep{van2017hybrid}.
\citet{wiering2008ensemble} gave a methodological overview over ensemble methods for integrating experience from multiple algorithms when forming policies.
Notably, Boltzmann addition and Boltzmann multiplication could integrate policies from multiple sources before action selection \citep{wiener1948cybernetics}.
Both \citet{wiering2008ensemble} and \citet{van2017hybrid} propose ways multiple off-policy learners could be involved in forming policy.
From these demonstrations on how multi-learner agents are possible, we shall dive further into the mechanism of behavior synthesis.
But first, some neuroscience.

\section{Neural Representation of Euclidean Space}
The 1906 Nobel price in physiology and medicine was awarded Santiago Ramón Y Cajal for work initiating the neuron doctrine \citep{cajal1911histologie}, 
    claiming that behavior originates from a network of cells with signaling capabilities rather than a monolithic soul. 
The neuron doctrine supplied a mechanistic understanding of biological computation as a distributed network of weak computational units.
Only by network phenomena and a delicately connected net of neurons can decisions, policies, and ultimately behavior emerge.
Eric Kandel later reported how synaptic connections change with use and how learning and memory are consequences of synaptic plasticity \citep{kandel1965heterosynaptic}.
Before the neuron doctrine, the consensus was that behavior and decision-making originate from a monolithic entity that followed us in this life and beyond -- \emph{the soul}.

Neural Representation of Euclidean Space (NRES) have been reported for different Euclidean spaces on a per-neuron cellular activation:
    when the Euclidean coordinate falls within the receptive field of an NRES neuron, the neuron fires with a heightened firing frequency.
A growing number of NRES modalities have been identified, with notable examples for navigation being
    place cells \citep{o1971hippocampus}, head-direction cells \citep{taube1990head}, and object-vector cells \citep{hoydal2020}.
While some NRES neurons have simple receptive fields centered around a coordinate, others have complicated repeating shapes like the hexagonal pattern of \emph{grid cells} \citep{moser2008place}.
For a comprehensive review of NRES modalities identified in neuroscience, see \citep{bicanski2020neuronal}.

Neural state is very different from the monolithic state of RL.
Analogous to separate cells representing coordinates of one Euclidean space, separate NRES modalities reflect different aspects of the navigational state.
The receptive fields of NRES neurons have a systematic increase from the dorsal to the ventral pole of the hippocampus \citep{fyhn2008grid, kjelstrup2008finite, solstad2009neural},
    allowing for NRES maps of multiple resolutions in parallel. 
The fully distributed representation of state thus allows for learning state representation 
    by individual receptive fields, for different NRES resolutions and across NRES modalities in parallel.
The monolithic Markov state of RL \citep{sutton2018reinforcement}, on the other hand, could explain difficulties for robot interaction learning \citep{kaelbling2020foundation}.
The most protruding difference between AI and neural state representation lies in the distributed nature of NRES.
We now explore how this can be emulated for RL systems.

%% file: src/methods.tex
\section{Decomposing the Prediction Problem} 


\begin{figure}[tb!] 
    \centering
    \fbox{\includegraphics[width=0.6\columnwidth]{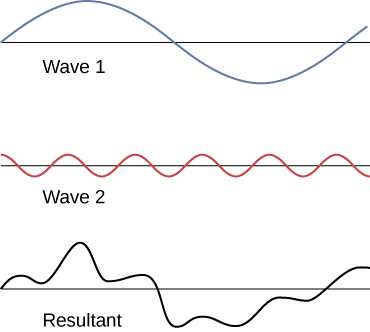}}
    \caption{Two simple sinusoidal functions can be combined to a complex function by superposition.
        \mbox{\citep{ling2016university}}
    }
    \label{figSuperpositionOpenstax}
\end{figure}

    The purpose of an \emph{agent} in reinforcement learning is to establish a proper behavior as defined by a reward signal.
    The agent improves behavior based on two intertwined aspects of experience:
    (1) The \emph{prediction problem} for learning the value of visiting states or state-action pairs as defined by the environment representation,
    and (2) The \emph{control problem} for selecting the most appropriate action based on the value as learned by the prediction problem.
    In this section, we expand on the concept of the prediction problem by considering the value function as a potential field across orthogonal reward signals.


    Let Orthogonal Value Functions (OVFs) be value functions of the state space $\mathbb{S}$ that adhere to mutually exclusive reward signals in $\mathbb{S}$.
    A relevant analogy would be to think of the value function as a potential field between different sources of energy.
    With multiple forces working on an object, the resultant work can be found as a linear combination of components.
    Similarly, a set of independent reward functions in $\mathbb{S}$ acting on agent value function can form a basis for agent value function in $\mathbb{S}$.
    NRES with mutually exclusive receptive fields is a good candidate for independent reward signals;
        with the place cell as our leading example, it is simple to visualize how agent position activates receptive fields and OVFs.
    %
    %
    %
    %
    %
    Each \emph{learner} has a simple reward shape, with a positive reward of $\mathbb{R}=+1$ upon activation of the corresponding NRES cell and $\mathbb{R}=0$ otherwise.
    A separate learner form the OVF according to reward signals as defined by mutually exclusive receptive fields of $\mathbb{S}$.
    %

    %

\begin{figure}[bt!h] 
    \centering
    \fbox{\includegraphics[width=0.6\columnwidth]{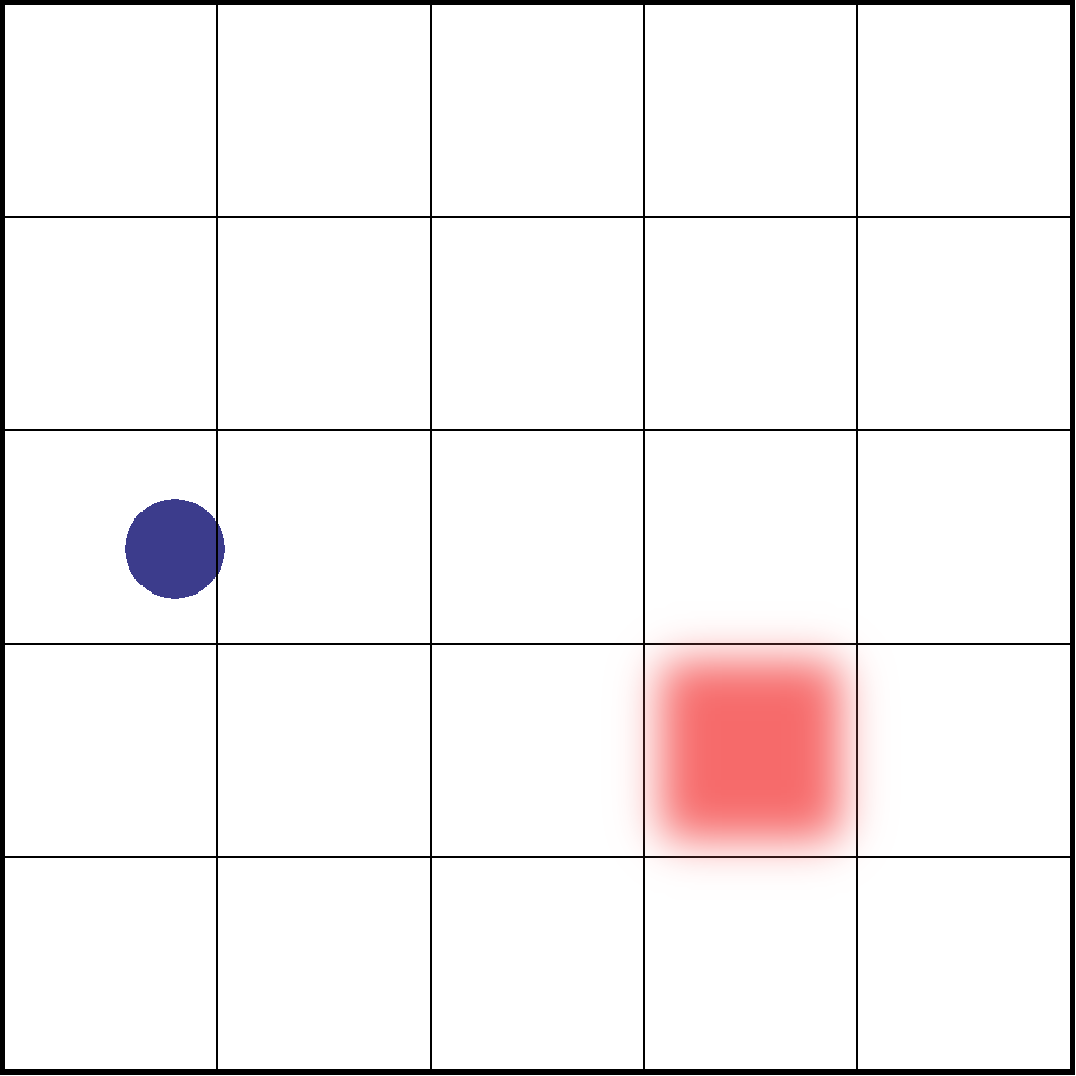}}
    \caption{
    An agent in $N5$ allocentric place-cell representation of Euclidean space:
    An $N5$ representation involves that each axis is divided into 5 equal intervals.
    A learner could, for example, form the OVF toward cell $(4, 4)$, with a reward signal defined by the activation of the corresponding NRES cell.
    The reward function of this particular learner is illustrated in red for feature $s_R\in\mathbb{S}$.
    The current parameter configuration of the agent defines from which $s\in\mathbb{S}$ this NRES modality's value function is extracted. 
%
    }
    \label{figLearningTheEnv_N5}
\end{figure}

Let there be $K$ individual learners, one for every receptive field of an NRES representation $\mathbb{S}$. 
With mutually exclusive receptive fields, the set of learners in $\mathbb{S}$ can be considered an orthogonal basis of the value function in this representation.
Value functions of $\mathbb{S}$ can be expressed as a linear combination of OVFs formed by the $K$ learners, allowing a neoRL agent to synthesize a range of behaviors.
The challenge of learning apt behavior now reduces to learning priorities between policies expressed via OVFs. 
Estimating scalar values based on supervised samples is a well-studied field in machine learning.
However, for the sake of clarity, static priorities defined by the associated reward is used.

\subsection{The Control Problem by Superposition} 
    The motivation for learning the value function is ultimately to form an effective policy for the challenge at hand.
    A simple challenge in Euclidean space can be for the agent to move to one particular position, activating feature $s_x$.
        If learners use Q-learning to establishing a potential that contributes to the \emph{Q-field} of the agent, the next action can be chosen by
        $$ a = \text{argmax}_a Q_{tot}(s, a) $$
        where $Q_{tot}$ is the resultant Q-field of the current situation. 
     With a single learner as input to the agent value potential,
     the agent's prediction problem becomes equivalent to that of the single learner, 
     and the mechanism surrounding the value function of the agent simplifies to that of a monolithic agent.

    For slightly more interesting challenges, multiple rewards can be expressed in the decomposed NRES representation.
    Each learner can be said to represent one consideration in this environment, learning the value function related to activating the corresponding NRES cell.
    When multiple considerations have priority, the superposition principle allows the Q-field to form over relevant OVFs. 
    \begin{equation}
      Q_{tot}(s, a) = \sum_{i\in\mathbb{S}_R} Q_{\mathcal{L}i}(s,a)
    \end{equation}
    where $\mathbb{S}_R$ is the set of NRES cells associated with reward and $Q_{\mathcal{L}i}(s,a)$ represent learner $\mathcal{L}_i$'s value component. 
    The $K$ learners in the full features set can thus be considered to be \emph{peer learners} for the task of navigating the environment representation.
    $$ \mathbb{S}_R = \{  s \in \mathbb{S} \,\; \Big| \quad |\mathbb{R}_s| > 0 \} $$

    An elegant approach would be to consider rewards to be linked to elements of interest in the environment rather than allocentric features:
    Let an \emph{Element of Interest} ($\xi_i$) be an instance in the environment associated with a reward.
    Assume for now that the priority and Euclidean parameter configuration of every element of interest in the set $\mathbb{E} = \{\xi_i\}$ is provided by the environment.
    Any parameter configuration is possible to map uniquely to the mutually exclusive NRES feature map $\mathbb{S}$.
    With element $i$'s importance $w_i$ proportional to the reward associated with the element activating feature $s$, the corresponding peer learner's contribution to the Q-field becomes:
    \begin{equation}
      \label{eqValueForOoI}
      Q_{tot}(s, a) = \sum_{i\in\mathbb{S}_R} w_i Q_{\mathcal{L}i}(s,a)
    \end{equation}
    Isolating rewards that comes from elements of interest, i.e. abstaining from utilizing timestep rewards or other shaped rewards, 
        the set of rewarded states is defined by the set of NRES cells occupied by an element of interest $\xi_i$. 
    \begin{equation}
      \label{eqRewardedStatesIsWhere_EiI_is}
        \mathbb{S}_R = \{ s \in \mathbb{S} \,\; \big| \; \exists \xi_i \in \mathbb{E}, \; \xi_i \in s \}  
    \end{equation}

    Note that an element of interest can be any element associated with a reward in a particular state set representation, decoupling the prediction problem in an environment from the rewards of one task.
    Experience expressed by distributed Q-fields is more general than monolithic value functions;
    In the neoRL approach, moving rewards or changing agent priorities during an agent's life-time does not require retraining the agent. 

    \begin{figure}[bt!h] 
        \centering
        \fbox{\includegraphics[width=0.6\columnwidth]{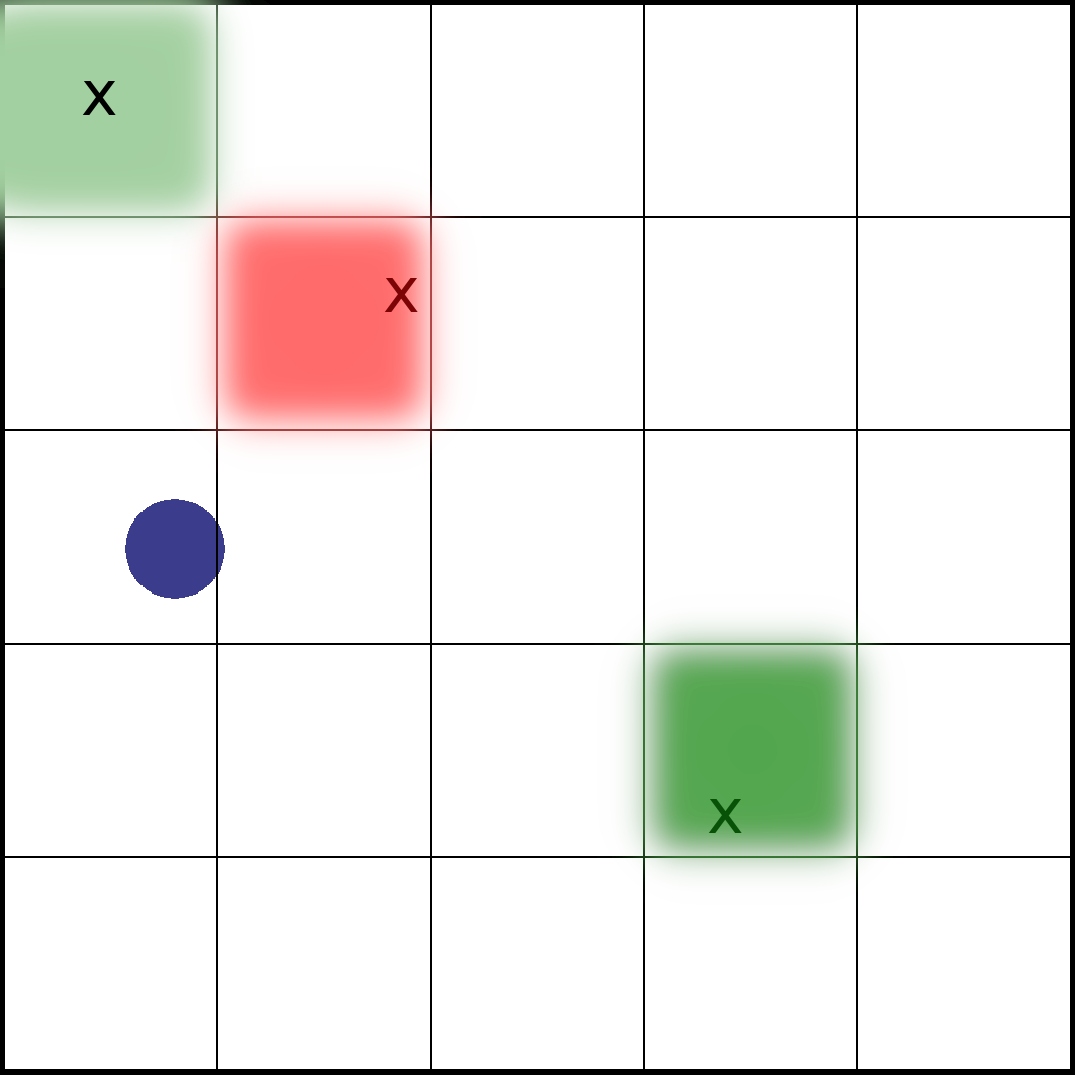}}
        \caption{
            Element of Interest (EoI) activates desires for allocentric features according to their importance:
            An EoI situated in feature $(4, 4)$ makes this desirable with $1.0$ ,
            another positive EoI activates feature $(1,1)$ with priority $0.5$ , 
            as represented by a green with lower saturation.
          An aversive element located in feature $(2,2)$
          activates the corresponding learner with a negative weight $w_i < 0$.
        }
        \label{figBluedotWithEoI}
    \end{figure}

%% file: src/experiments.tex
\section{Experiments} 

Algorithms in RL learn behavior by interaction with the environment, making the environment defining for the outcome of any RL experiment. 
Numerous environments exist to highlight challenges for state-of-the-art reinforcement learning agents.
%
Learning autonomous navigation in allocentric space does not seem to get much attention, as finding appropriate test-environments can be difficult.
%
Preferably, an environment for autonomous real-world navigation learning is represented by continuous allocentric coordinates and with a complexity that requires reactive navigation.
Real-time execution would be a plus since it limits the amount of training data available to the agent to a realistic order of magnitude. 
Physical systems generally depend on temporal aspects like inertia. 
Most of these qualities can be found in Karpathy's WaterWorld challenge.

\section{WaterWorld}
\label{sec:experiments}

\begin{figure}[tb!]
    \centering
    \fbox{\includegraphics[width=0.3\textwidth]{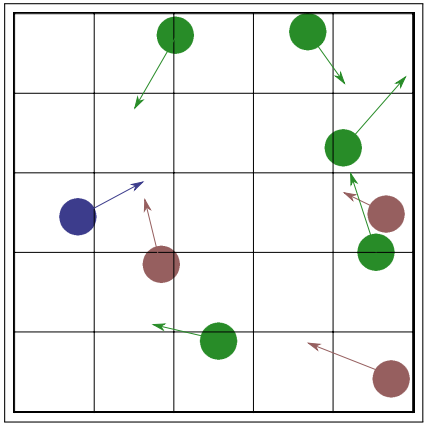}}
    \caption{NRES $N5$ representation of Element-of-Interest (EoI) in the \emph{WaterWorld} environment.
    Each EoI and the location of the agent represented in the PlaceCell NRES modality.
    Red and Green represent the demeanor of each creep, whereas Blue represents the current agent location.
    In addition, arrows have been drawn to illustrate the current speed vector of each element.
    }
    \label{figDiscretizationN5}
\end{figure}

Karpathy's WaterWorld challenge as implemented in Pygame learning environment(PLE) \citep{tasfi2016PLE} is an environment with a continuous 2D resolution, 
    inertia dynamics and external considerations referred to as \emph{creeps}.
Creeps move with a constant speed vector, reflected when hitting a wall.
Creeps have a demeanor, as illustrated by color: green creeps are desirable with [+1] reward, and red creeps are repulsive with [-1] reward upon capture.
When the agent captures a creep, a new one is initialized with a random speed, position, and demeanor -- causing a chaotic scenario that requires reactive navigation.
When all green creeps have been captured, the board is restarted with an accompanying [+5] reward.
In all experiments, a constant number of $8$ creeps have been used, as illustrated in Figure \ref{figDiscretizationN5}.
We find the allocentric PyGame implementation \citep{tasfi2016PLE} of WaterWorld appropriate for RL research for real-time navigation autonomy.
However, the environment is listed as unsolved \citep{webWaterWorldUnsolvedOpenai} -- making comparisons to alternative solutions difficult.

Instantaneous information regarding elements-of-interest (EoI), i.e., the position and demeanor of each creep, is provided by the environment.
Demeanor defines the reward associated with the creep, crucial for priority $w_i$ associated with EoI $i$ by equation \ref{eqValueForOoI}.
Positions are represented in 2D allocentric coordinates from the environment, allowing for extracting $\xi_i \in \mathbb{S}$ for the Place Cell NRES modality of EoI $i$.
Basal actions affect the agent by accelerating it in the cardinal directions, [N, S, E, W]. 


\subsubsection{Allocentric Position Modality, Single layer:}
Our primary assumption is that the agent value function in effect can be considered a potential field across OVFs, pulling the agent toward the next decision.
Our first experiment explores to what degree the superposition principle holds for the value function of individual learners. 
We compare the accumulated score of neoRL agents based on single-res NRES to Brownian motion, i.e., an $\epsilon$-greedy policy with $\epsilon = 1.0$.
Under the convention used in Figure \ref{figDiscretizationN5}, where $N5$ signifies an NRES map with $5x5$ tiles, 
    five different resolutions are explored from $N10$ to $N90$.
All experiments were conducted over 150.000 time-steps for each neoRL agent. 

\subsubsection{Allocentric Position Modality, Multiple resolutions: }
\begin{figure}[bt!h]
    \centering
    \includegraphics[width=1.05\columnwidth]{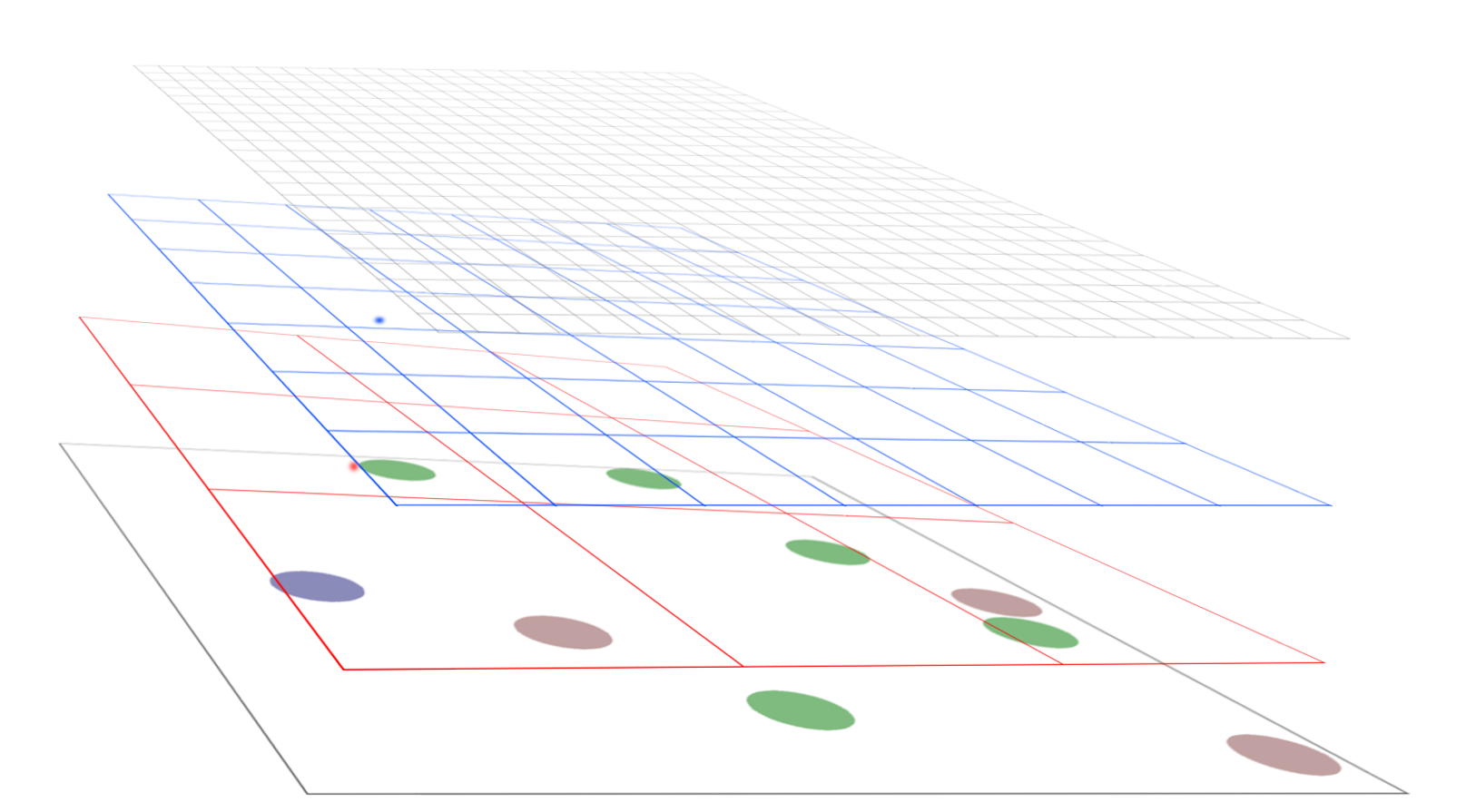}
    \caption{Illustration of multiple state representation in the decision
    agent, where each tile represent the objective of its respective learner.
    [Red] N3 representation 
    [Blue] N7 representation
    [Black] N23 representation. 
    }
    \label{figMultiShepherdIllustration}
\end{figure}

Our second experiment explores how integrating experience across multiple state spaces affect neoRL performance.
%
An interpretation of the progressive increase for receptive fields in the ventral direction of the hippocampus is that different NRES maps exist with different resolutions. 
We adopt this view in experiment 2, where we let the neoRL agent combine value function across multiple NRES state representations.
In this experiment we assess whether the neoRL agent is capable of forming apt policies by integrating experience across multiple state spaces.
We compare the proficiency of a multi-res neoRL agent that learns over \{$N3$, $N7$, $N23$\} NRES state spaces to three single-res agents by $N3$, $N7$, and $N23$ NRES. 
The neoRL agent layout is illustrated in Figure \ref{figMultiShepherdIllustration}.
Prime numbers are used as the resolution for each layer, minimizing the potential for overlapping boundaries. 
The resulting $587$ learners in the multi-res agent learn in parallel by off-policy learning.
In this setup, the contribution of each learner is inversely proportional to the size of its receptive field.

One approach of measuring the proficiency of the agent is as the per-timestep average reward across parallel runs.
We are interested in real-time learning efficiency and initialize a neoRL agent with no priors at the beginning of each run.
%
A per-timestep average across $100$ independent runs provides information about the transient timecourse in navigation capabilities.
Note that every run starts with a separate neoRL agent with no prior experience.
All experiments are conducted on an average desktop computer, with one run taking somewhat under one hour on a single CPU core.

\subsection{Results}
Results are reported as real-time execution of agents as they learn, without any previous experience at the task.
Reported resolution for each experiment adheres to the convention from Figure \ref{figLearningTheEnv_N5}, dividing each axis of the Euclidean space into $N$ steps. 
The x-axis of all plots represents the number of time steps since the beginning of a run, i.e., the real-time execution in time-steps since initiation of the agent.

\subsubsection{Allocentric Position Modality, Single layer}
\begin{figure*}[tbp]
    \centering
    \includegraphics[width=0.82\textwidth]{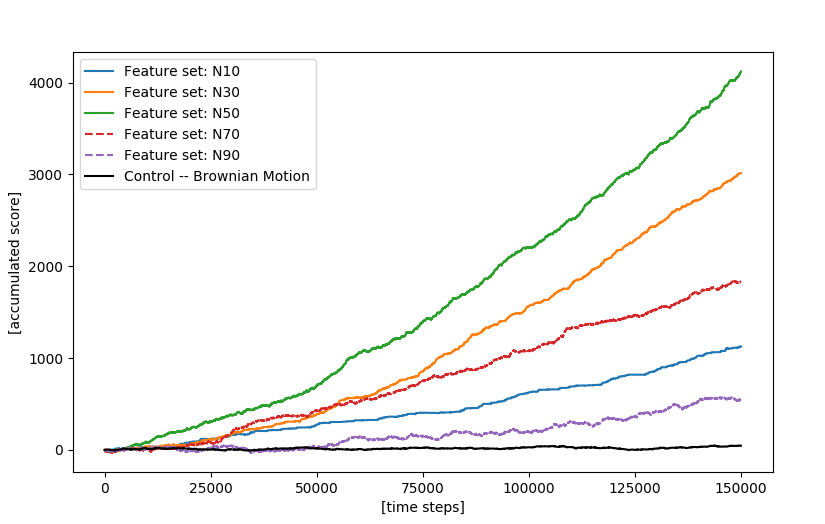}
    \caption{Accumulated Reward by peer agents with elements of interest 
    for runs with grid coding resolutions, $N10 - N90$ over $150.000$ time steps.
    Brownian motion in black is believed to be comparable to a first run of an untrained Deep RL agent.
    }
    \label{figShepherdsN5_N90}
\end{figure*}

A distributed representation of the Markov state is plausible for neoRL agents.
Figure \ref{figShepherdsN5_N90} shows the accumulated score of neoRL agents with NRES Place Cell representations from $N10$ to $N90$.
All neoRL agents perform better than control.
Brownian motion seems incapable of achieving a single board reset since the accumulated score fluctuates around 0 for the length of the experiment. 
All neoRL agents are capable of accumulating a significant amount of experience, verifying that OVF can function as a basis for synthezing successful behavior.

A strong correlation between NRES resolution and proficiency at the task can also be observed in Figure \ref{figShepherdsN5_N90}.
The immediate proficiency at the task can be seen from the steepness of the curve.
Agents based on lower NRES resolution initially learn quicker than agents with higher NRES resolution.
However, neoRL agents based on lower NRES resolutions seem to saturate at a lower proficiency.
For these particular runs, with $8$ creeps and during a $150.000$ time step interval, the $N50$ representation appears to achieve the highest score.
Although this number is task-specific, it is worth noting how all neoRL agents are comparable in learning speed.
Despite $N70$ NRES having almost $50$ times the dimensionality of $N10$\footnote{The $N10$ representation is comprised of $100$ receptive fields, whereas the finer $N70$ resolutions have $4900$ receptive fields.},
    the two neoRL agents based on these representations are comparable in learning.
This effect requires further attention.



\begin{figure*}[tbp]
    \centering
    \includegraphics[width=0.82\textwidth]{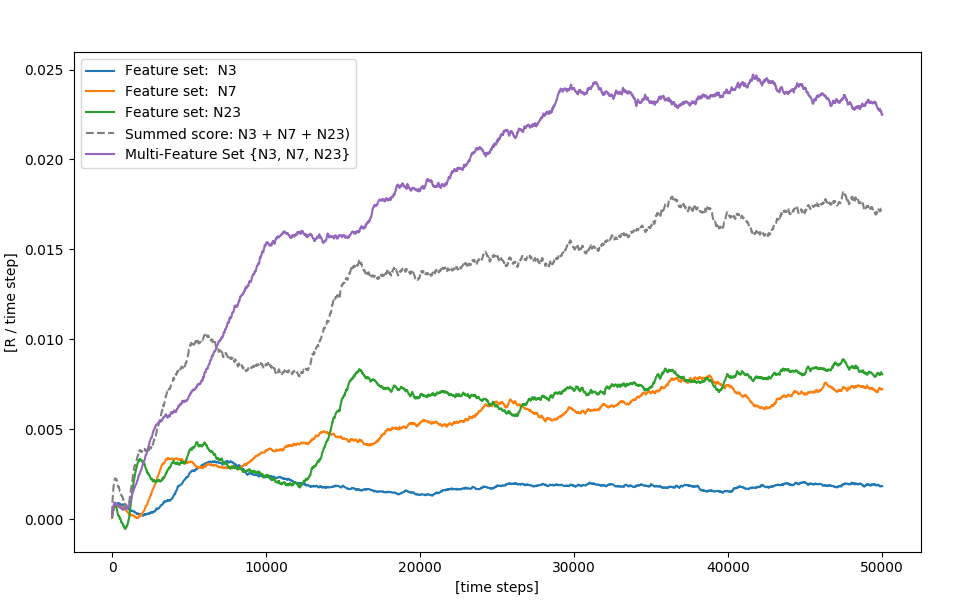}
    \caption{
        The neoRL agent is capable of incorporating experience from multiple state sets for navigation. 
        A neoRL agent with experience from all three layers seen in Fig. \ref{figMultiShepherdIllustration} (purple) 
            performs better than neoRL agents based on the individual NRES layer (blue, orange, green).
        The grey line represents the algebraic sum of the mono-res agents, highlighting that the multi-res neoRL agent performs better than the sum of its parts.
        Each curve is a presentation of the per-timestep average of $100$ independent runs. 
        %
        }
    \label{figInterdimensionalLearningEfficiency}
\end{figure*}

\subsubsection{Allocentric Position Modality, Multiple resolutions}
Combining the value potential from multiple representations of state can significantly increase navigation performance.
The transient proficiency of the neoRL agent in the four experiments, $N3$, $N7$, $N23$, and multi-res \{$N3$, $N7$, $N23$\}, is presented in Figure \ref{figInterdimensionalLearningEfficiency}.
Each curve is the result of a per-timestep average over $100$ independent runs. 
These results verify without any doubt that neoRL agents benefit from combining experience across multiple NRES feature sets.
With the algebraic sum of the per-timestep proficiency of the three mono-res agents shown in grey, 
    we see that the multi-res neoRL agent learns quicker, to higher proficiency, than the sum of its parts.

The superposition principle for behavior across state spaces seems to alleviate the curse of dimensionality:
The almost 6-fold increase in the number of states (from $7^2 = 49$ to $3^2+7^2+23^2 = 290$ states) resulted in a 3.5-factor increase in received reward without increasing training time.
Figure \ref{figInterdimensionalLearningEfficiency} shows that learning happens as fast or possibly a little faster for the multi-res agent than for the $N7$ mono-res agent.
This effect could be defining for real-world interaction learning and requires further attention.

%% file: src/discussion.tex
\section{Discussion}


    Navigation autonomy is plausible in real-time by RL agents with an emulated neural representation of space. 
    NRES-Oriented RL (neoRL) agents are possible due to developed theory on orthogonality in the value domain, allowing for behavior synthesis across multiple learners.

    Whereas neural systems are capable of autonomous navigation, modern technology is not.
    The most protruding difference between these systems is how state is represented. 
    Digital RL systems require a monolithic state concept, whereas neural systems work by patterns of activation.
    The Markov state in RL holds enough information to uniquely define the probability distribution of the next state \citep{sutton2018reinforcement}.
    The Markov decision process works well with deep function approximation, and RL agents supported by deep learning have mastered a selection of board games.
    However, deep RL agents require much training, do not generalize, and are neither incremental nor compositional \citep{kaelbling2020foundation}.
    With deep RL appearing to struggle with real-world interaction learning, we have looked elsewhere for inspiration.
    Evidence suggests that Neural Representation of Euclidean Space (NRES) represent Euclidean coordinates by activation patterns on the per-neuron level.
    An NRES set $\mathbb{S}$ with mutually exclusive receptive fields provides a set of orthogonal reward signals of $\mathbb{S}$.
    Utilizing these signals as reward signal for independent learners, the set of Orthogonal Value Functions (OVFs) form a basis for any reward function of $\mathbb{S}$.
    Experiments verify that NRES-Oriented RL (neoRL) agents are capable of forming skilled navigation while learning.

    This study can be considered a plausibility study on \mbox{neoRL} capabilities for real-time interaction learning in Euclidean spaces.
    An important continuation would be to perform a thorough mathematical analysis on the matter of orthogonality.
    Specifically, deriving the equations for how singular reward functions cause orthogonal value functions can cause a better understanding of behavior synthesis.
    In experiment 2, we have seen how different state-space representations of the same parameter set can improve performance.
    We believe the same to be possible for state spaces across different parameter spaces, but this has not been tried.
    Secondly, the priority $w_i$ in Equation \ref{eqValueForOoI} allows for weighing the OVF with the reward associated with the element-of-interest.
    In this work, $w_i$ has remained static for the sake of clarity.
    The agent could also learn the association between element $i$ and global reward $\mathbb{R}$, but this remains for further research.
    Lastly, all experiments conducted on the neoRL framework have been on the WaterWorld environment.
    This is a great environment for navigation. However, it would be interesting to explore other challenges.
    These challenges could be across different Euclidean spaces, e.g., the space comprised of joint-angles for a robotic arm.